\documentclass[journal,twoside,web]{ieeecolor}
\usepackage{tmi}
\usepackage{cite}
\usepackage{amsmath,amssymb,amsfonts}
\usepackage{bm}
\usepackage{algorithmic}
\usepackage{graphicx}
\usepackage{textcomp}
\usepackage{orcidlink} %

\usepackage{tabularx, booktabs}
\usepackage{multirow, multicol}
\hypersetup{
            colorlinks=true,
            linkcolor=blue,
            anchorcolor=blue,
            citecolor=blue
            }

\def\BibTeX{{\rm B\kern-.05em{\sc i\kern-.025em b}\kern-.08em
    T\kern-.1667em\lower.7ex\hbox{E}\kern-.125emX}}
\begin{document}
\title{Cross-Modal Clustering-Guided Negative Sampling for Self-Supervised Joint Learning from Medical Images and Reports}
\author{Libin Lan$^{\orcidlink{0000-0003-4754-813X}}$, \IEEEmembership{Member, IEEE}, Hongxing Li$^{\orcidlink{0009-0002-7958-3976}}$, Zunhui Xia$^{\orcidlink{0009-0008-6706-5817}}$, Juan Zhou$^{\orcidlink{0009-0008-0243-3949}}$, Xiaofei Zhu$^{\orcidlink{0000-0001-8239-7176}}$, Yongmei Li$^{\orcidlink{0000-0003-2829-6416}}$, Yudong Zhang$^{\orcidlink{0000-0002-4870-1493}}$, \IEEEmembership{Senior Member, IEEE}, and Xin Luo$^{\orcidlink{0000-0002-1348-5305}}$, \IEEEmembership{Fellow, IEEE}
\thanks{This work described in this paper was supported in part by the Scientific Research Foundation of Chongqing University of Technology under Grant 2021ZDZ030, in part by the Youth Project of Science and Technology Research Program of Chongqing Education Commission of China under Grant KJQN202301145, and in part by the National Natural Science Foundation of China under Grant 62472059. \textit{(Corresponding authors: Xin Luo.)} }
\thanks{Libin Lan, Hongxing Li, Zunhui Xia, and Xiaofei Zhu are with the College of Computer Science and Engineering, Chongqing University of Technology, Chongqing 400054, China (e-mail: lanlbn@cqut.edu.cn; hongxing.li@stu.cqut.edu.cn; zunhui.xia@stu.cqut.edu.cn; zxf@cqut.edu.cn).}
\thanks{Juan Zhou is with the Department of Pharmacy, the Second Affiliated Hospital of Army Military Medical University, Chongqing 400037, China (e-mail: zhoujuan0523@tmmu.edu.cn).}
\thanks{Yongmei Li is with the Department of Radiology, the First Affiliated Hospital of Chongqing Medical University, Chongqing 400016, China (e-mail: lymzhang70@aliyun.com). }
\thanks{Yudong Zhang is with the School of Computer Science and Engineering, Southeast University, Nanjing, Jiangsu 210096, China (e-mail: yudongzhang@ieee.org).}
\thanks{Xin Luo is with the College of Computer and Information Science, Southwest University, Chongqing 400715, China (e-mail: luoxin@swu.edu.cn).}
}

\maketitle

\begin{abstract}

Learning medical visual representations directly from paired images and reports through multimodal self-supervised learning has emerged as a novel and efficient approach to digital diagnosis in recent years. However, existing models suffer from several severe limitations. 1) neglecting the selection of negative samples, resulting in the scarcity of hard negatives and the inclusion of false negatives; 2) focusing on global feature extraction, but overlooking the fine-grained local details that are crucial for medical image recognition tasks; and 3) contrastive learning primarily targets high-level features but ignoring low-level details which are essential for accurate medical analysis. Motivated by these critical issues, this paper presents a Cross-Modal Cluster-Guided Negative Sampling (CM-CGNS) method with two-fold ideas. First, it extends the k-means clustering used for local text features in the single-modal domain to the multimodal domain through cross-modal attention. This improvement increases the number of negative samples and boosts the model representation capability. Second, it introduces a Cross-Modal Masked Image Reconstruction (CM-MIR) module that leverages local text-to-image features obtained via cross-modal attention to reconstruct masked local image regions. This module significantly strengthens the model's cross-modal information interaction capabilities and retains low-level image features essential for downstream tasks. By well handling the aforementioned limitations, the proposed CM-CGNS can learn effective and robust medical visual representations suitable for various recognition tasks. Extensive experimental results on classification, detection, and segmentation tasks across five downstream datasets show that our method outperforms state-of-the-art approaches on multiple metrics, verifying its superior performance. Our code is available at \url{https://github.com/violet-42/CM-CGNS}.

\end{abstract}

\begin{IEEEkeywords}
Clustering-guided negative sampling, contrastive learning, masked image reconstruction, multimodal self-supervised learning.
\end{IEEEkeywords}

\section{Introduction}
\label{sec:introduction}
\IEEEPARstart{I}{n} the domain of natural images, the use of large-scale labeled datasets \cite{deng2009imagenet} has significantly propelled the advancement of deep learning, leading to remarkable achievements in visual recognition \cite{he2016deep},\cite{dosovitskiy2021an}. However, in medical imaging, obtaining high-quality manually annotated datasets poses a considerable challenge due to the requirement for annotations by an experienced physician, which is both time-consuming and expensive. Therefore, the development of deep learning in the field of medical imaging has been somewhat constrained. In the context of natural images, to address the issue of insufficient labeled datasets, pre-training methods are commonly adopted, where models are first pre-trained on large-scale natural image datasets, and then the learned general image representations are transferred to downstream tasks, thereby enhancing the convergence speed and generalization performance of the downstream models. However, due to the domain differences between natural and medical images, directly transferring models pre-trained on natural image datasets like ImageNet \cite{deng2009imagenet} to downstream medical imaging tasks often yields suboptimal results. To tackle this issue, a mainstream approach \cite{zhou2023self}, \cite{tang2022self} is to employ self-supervised methods for pre-training on large-scale unlabeled medical image datasets. Nevertheless, this approach still faces two primary challenges. First, the amount of medical image data available for pre-training is significantly less compared to the volume of natural image data. Second, medical images place greater emphasis on local features and fine-grained information within the images, as opposed to natural images.

\begin{figure*}
\centering
\includegraphics[width=1.0\linewidth, keepaspectratio]{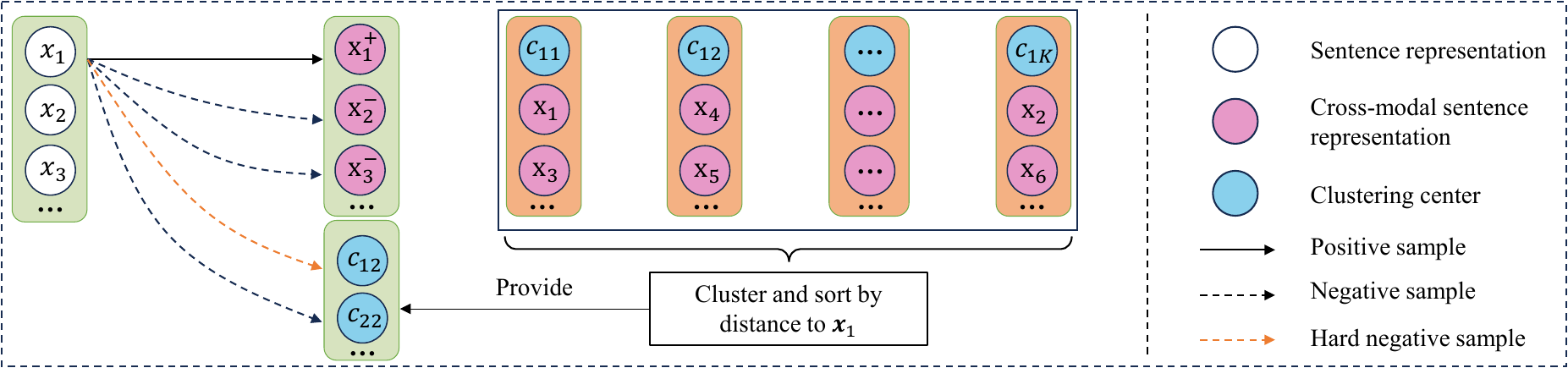}
\caption{Illustrations of our proposed Cross-Modal Clustering-Guided Negative Sampling. Compared to previous contrastive learning, our method first clusters the cross-modal sentence representations, resulting in $K$ clusters and their respective centers. Then, we sort the cluster centers in descending order based on the distances between the cluster centers and a sentence representation (e.g., $x_1$), yielding $(c_{11}, c_{12}, ..., c_{1K})$. For $x_1$, we take $c_{12}$ as its corresponding hard negative. Similarly, we can derive the hard negative $c_{i2}$ for sentence representation $x_i$. When calculating the contrastive loss, all the hard negatives are added to the negative sample queue, thus increasing the number of negative samples. Additionally, by assigning a greater weight\iffalse to the hard negative $c_{i2}$ corresponding to the sample $x_i$\fi, the performance of contrastive learning can be further enhanced.}
\label{fig1:cluster}
\end{figure*}

To address the issue of limited medical image data, a natural idea is to leverage paired radiology reports to learn transferable medical visual representations through Vision-Language Pre-training (VLP) \cite{li2021align}, \cite{radford2021learning}.
Radiology reports are written by medical experts such as radiologists in clinical practice, thus providing a valuable source of semantic information that offers supervision information for the images. This can help offset the scarcity of image data and the high cost of annotations, thereby assisting the model in learning richer and deeper feature representations of medical images. In recent years, VLP has seen significant advancements in the medical field, particularly through the design of reasonable pretext tasks, such as contrastive learning \cite{he2020momentum}, \cite{wang2022medclip} and generative reconstruction tasks \cite{zhang2023generalized},\cite{li2023harnessing}. However, most contrastive learning methods are designed from a global perspective and do not take into account the detailed information within images and texts. While the generated representations are well-suited for image-level tasks, such as image classification, they are not ideal for pixel-level tasks that require fine-grained details, such as object detection and semantic segmentation.

Some works have considered local detail information. GloRIA \cite{huang2021gloria} adopts an attention-based contrastive learning strategy to align sub-region image representations with word-level representations in reports, enabling the model to focus more on specific parts of the images and key terms in the text. Similarly, MGCA \cite{wang2022multi} develops a bidirectional attention-based token-wise alignment model, further refining the correspondence between images and texts. AdaMatch\cite{chen2024fine} introduces the Adaptive Patch Extraction (AdaPatch) module, which can adaptively select image regions based on the actual size and location of lesions, thereby capturing lesion characteristics more accurately. However, while these word-level local alignment strategies enhance the model's ability to understand local details, they often overlook the semantic structure information at the sentence level in medical reports. Medical reports typically contain complex syntactic and semantic relationships, and the associations between words are crucial for comprehending the overall meaning of the report. 
LOVT \cite{muller2022joint} and PRIOR \cite{cheng2023prior}, although utilizing sentence-level semantic information for contrastive learning, only draw negative samples from sentences within the same sample, leading to insufficient numbers of negative samples and limiting model performance.

In this paper, we focus on joint representation learning from paired medical images and radiology reports. We propose a novel approach, \textbf{C}ross-\textbf{M}odal \textbf{C}lustering-\textbf{G}uided \textbf{N}egative \textbf{S}ampling (\textbf{CM-CGNS}), which integrates contrastive learning with clustering algorithms to seamlessly leverage both global and local features from medical images and radiology texts. This method facilitates more generalized and robust medical visual representation learning. We treat image sub-regions and text sentences as the basic units for conducting both global and local representation contrastive learning. Global image and global text features are derived by respectively applying attention pooling to the local image and local text features extracted by the encoders, and these global features are subsequently used to compute the global contrastive loss. A cross-modal clustering algorithm, which uses local image-text features obtained through cross-modal attention, is employed to generate additional negative samples. These generated negative samples are then used to compute the local contrastive loss.
The incorporation of multimodal clustering significantly boosts overall model performance, as shown in Fig. \ref{fig1:cluster}. 
Additionally, to learn more accurate local features and finer pixel-level details, masked image reconstruction is essential. To this end, we design a cross-modal masked image reconstruction module that leverages cross-modal text-to-image information to reconstruct masked image regions.

Our main contributions are three-fold:
\begin{itemize}
  \item[•] We present the Cross-Modal Clustering-Guided Negative Sampling (CM-CGNS) method, which leverages clustering algorithms toward cross-modal data to expand the number of negative samples required for contrastive learning. By addressing a critical limitation in existing models -- namely, the scarcity of hard negative samples and the inclusion of false negatives -- CM-CGNS achieves superior cross-modal local representation alignment.
  \item[•] We propose the Cross-Modal Masked Image Reconstruction (CM-MIR) module, which reconstructs masked image regions using cross-modal local text-to-image information. CM-MIR facilitates more effective fusion of textual and visual data and preserves low-level image features essential for downstream medical image recognition tasks.
  \item[•] CM-CGNS outperforms existing state-of-the-art methods on three basic visual recognition tasks: supervised classification, object detection, and semantic segmentation, achieving superior performance under almost all evaluation metrics.
  \end{itemize}
  
The remainder of this paper is organized as follows. Section \ref{sec:relatedwork} delves into previous relevant work and its limitations. Section \ref{sec:method} details our model architecture and the innovations of our method. Section \ref{sec:experiments} presents the experimental details and results including the visualization of the cross-modal attention maps, and provides an in-depth analysis of the results. Section \ref{sec:conclusion} summarizes our findings, points out the limitations of the work, and indicates directions for future research.

\section{Related Work}
\label{sec:relatedwork}
\subsection{Medical Vision-Language Pre-training}
Vision-Language Pre-training (VLP) \cite{zhou2020unified} aims to learn the semantic correspondence between visual and linguistic modalities using large-scale, unlabeled data through self-supervised pre-training.
Existing VLP methods such as ALBUF \cite{li2021align}, CLIP \cite{radford2021learning} have made significant advances in a variety of downstream recognition tasks within the natural image domain. Concurrently, the application of VLP in the medical field is becoming increasingly widespread due to its ability to extract generic vision-language representations from medical images and texts. Several notable works like Med-VLP \cite{chen2022align}, GLoRIA \cite{huang2021gloria}, LoVT \cite{muller2022joint}, MGCA \cite{wang2022multi}, and PRIOR \cite{cheng2023prior} have achieved remarkable success across a range of downstream recognition tasks in the medical visual domain. 
These works in both domains typically train models using unlabeled data through well-designed pretext tasks, which enable the models to learn effective and robust feature representations without the need for extensive labeled datasets.
According to \cite{gui2024survey}, pretext tasks generally fall into four main categories: context-based methods, contrastive learning, generative algorithms, and contrastive-generative methods. Among these, contrastive learning and generative approaches are most relevant to our work. In the following sections, we will discuss related studies concerning these two types of pretext tasks. 

\subsection{Contrastive Learning}
\subsubsection{ Contrastive Learning with Positive and Negative Sample Pairs}
Contrastive learning stands as one of the most extensively utilized methods in self-supervised learning. Its core thought revolves around learning the intrinsic features and structure of data by comparing similarities and differences between data samples. Specifically, it aims to make the embeddings of positive sample pairs as close as possible in the latent space while maximizing the distance between negative sample pairs. This process facilitates the model's ability to capture inherent patterns and relationships within the data. SimCLR \cite{chen2020simple} employs straightforward data augmentation strategies to generate positive sample pairs and treats other samples within the same batch as negative samples, thereby establishing an efficient contrastive learning framework. MoCo \cite{he2020momentum} introduces a dynamic queue mechanism for storing negative samples and uses a momentum update mechanism to maintain a large pool of negative samples, addressing the issue of limited negative sample numbers. 

\subsubsection{Contrastive Learning without Explicit Positive and Negative Sample Pairs}
Moving beyond the traditional approach of constructing positive and negative sample pairs, recent years have seen the emergence of innovative contrastive learning methods that do not require explicit construction of such pairs. Representative works in this area involve SimSiam\cite{chen2021exploring}, BarlowTwins\cite{zbontar2021barlow}, and VICReg\cite{bardes2022vicreg}. These methods focus on directly optimizing specific metrics or objective functions rather than explicitly distinguishing between positive and negative samples. By preserving certain statistical properties between inputs -- such as invariance, variance maximization, or covariance minimization -- these approaches naturally prevent the model from collapsing into trivial solutions. Moreover, these approaches do not necessitate the construction of negative sample pairs, thus reducing the need for large batch sizes during training.

\subsubsection{Contrastive Learning Based on Clustering}
Recently, clustering-based contrastive learning has demonstrated significant potential in unsupervised representation learning. By employing online clustering methods, contrastive learning can be simultaneously performed  at both the instance level and the cluster level within the feature space, further enhancing the model learning capabilities. Contrastive Clustering (CC) \cite{li2021contrastive} and Twin Contrastive Learning (TCL) \cite{li2022twin} are prime examples of this approach, each proposing unique data augmentation strategies to promote more effective contrastive learning. ClusterNS \cite{deng2023clustering} integrates clustering information into contrastive learning to improve the negative sampling process in unsupervised sentence representation learning. Other works \cite{zhong2021graph}, \cite{liu2022deep} explore combining contrastive learning with graph clustering methods to enhance the discriminative power of feature representations while optimizing cluster assignments. However, these methods are typically trained on single-modal data, limiting their ability to fully leverage multimodal data for self-supervised learning. To address this issue, we propose the CM-CGNS method for clustering, which expands the number of negative samples used in contrastive learning. 

\subsection{Masked Image Modeling}
Masked Image Modeling (MIM) represents a class of generative algorithms that learn robust image representations by randomly masking portions of an input image and training the model to predict these masked regions. BEiT \cite{bao2022beit} introduces a tailored MIM task for visual pre-training using two-stage training, where the input image is decomposed into visual tokens, and a subset of these tokens is randomly masked. Conversely, MAE \cite{he2022masked} addresses the problem through one-stage training from the perspective of image signal sparsity without using visual tokens. SimMIM\cite{xie2022simmim}, on the other hand, provides a simplified MIM framework that directly masks pixels and reconstructs the masked regions using a decoder. However, these methods that rely solely on the image modality for masked region reconstruction fail to fully leverage multimodal information. To address this limitation, we propose CM-MIR, which utilizes cross-modal textual information to reconstruct masked image regions. This proposed approach enhances the model's ability to integrate multimodal information while preserving low-level image features.

\subsection {Medical Image and Report Representation Alignment}
Compared to natural images, medical images require a focus on fine-grained pathological details. Aligning features solely from a global perspective, as in models like CLIP \cite{radford2021learning}, falls short for medical data. Consequently, Med-VLP \cite{chen2022align} necessitates the development of specialized local modules to capture detailed pathological information. Recent works such as GLoRIA\cite{huang2021gloria}, LoVT\cite{muller2022joint}, MGCA\cite{wang2022multi}, and PRIOR \cite{cheng2023prior} have designed their own local alignment modules. Specifically, GLoRIA and MGCA align text words with image sub-regions, while LoVT and PRIOR align text sentences with image sub-regions. However, these methods mentioned above limit the number of negatives by sampling them only within the same text, which reduces the effectiveness of contrastive learning. To address this issue, we propose integrating local and global alignment into CM-CGNS to expand the range of the negative sample from within a single text to encompass the entire mini-batch, thus increasing the number of negative samples used in contrastive learning. Additionally, we assign greater weight using a grid search strategy to the obtained hard negatives, further enhancing the performance of contrastive learning. To further leverage local features, our proposed CM-MIR enhances the ability to fuse multimodal features while preserving low-level features beneficial for medical images.

\section{Method}
\label{sec:method}
\subsection{Overview}
\begin{figure*}[htbp]
\centering
\includegraphics[width=1.0\linewidth, keepaspectratio]{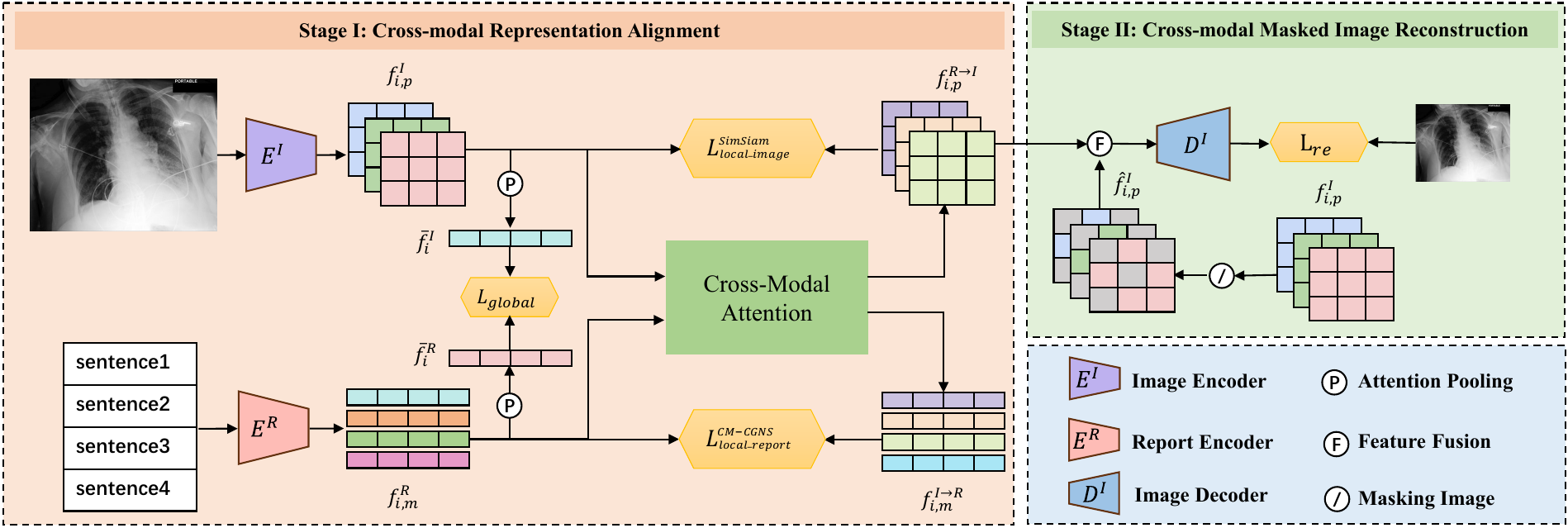}
\caption{The overall architecture of our proposed CM-CGNS. In Stage 1, for the input image-text pairs, we use an image encoder and a text encoder to extract local image features $f_{i,p}^{I}$ and local text features $f_{i,m}^{R}$, respectively. These local features are then aggregated into global image and text features through attention pooling. The alignment of global image-text pairs is achieved using a global contrastive loss $L_{global}$. For the alignment of local image-text pairs, we obtain cross-modal representations of images and texts through a cross-modal attention mechanism. We then compute $L_{local\_image}$ and $L_{local\_report}$ using the CM-CGNS and SimSiam \cite{chen2021exploring} methods, respectively, to align these local features. In Stage 2, we use the cross-modal image feature $f_{i,p}^{R\rightarrow I}$ to complete the masked image feature $f_{i,p}^{I}$, and calculate the image reconstruction loss $L_{re}$ through the cross-modal image reconstruction module.}
\label{fig2:framework}
\end{figure*}

The main objective of this work is to learn visual representations from medical images and paired radiology reports in a self-supervised learning manner, which are beneficial for various downstream image recognition tasks. Our overall architecture is illustrated in Fig. \ref{fig2:framework}. 
The model takes as input a set of data $\bm{S}=\{(\bm{x}_{1}^{I}, \bm{x}_{1}^{R}), (\bm{x}_{2}^{I}, \bm{x}_{2}^{R}), ..., (\bm{x}_{N}^{I}, \bm{x}_{N}^{R})\}$ consisting of $N$ pairs of images and reports, which are mapped into the latent space regarding local features yielded by an image encoder $E^I$ and a text encoder $E^R$. Subsequently, global features are extracted through attention pooling operation and applied to align both image and text representations.
To better achieve semantic alignment between images and texts, apart from standard global contrastive learning, we also employ local contrastive learning to capture the fine-grained image information required for downstream image tasks. 

Additionally, By cross-modal clustering-guided algorithms, we increase the number of negative samples used for sentence-level alignment and give hard negatives greater weight to enhance the model learning efficiency and representation capabilities. 
Furthermore, we design a cross-modal image reconstruction module to further leverage fine-grained image information. This module uses the local text-to-image features obtained via cross-attention to reconstruct masked local image features, thereby improving the model's cross-modal information interaction capabilities while preserving lower-level image features beneficial for downstream fine-grained tasks.

\begin{figure}
\centering
\includegraphics[width=0.95\linewidth, keepaspectratio]{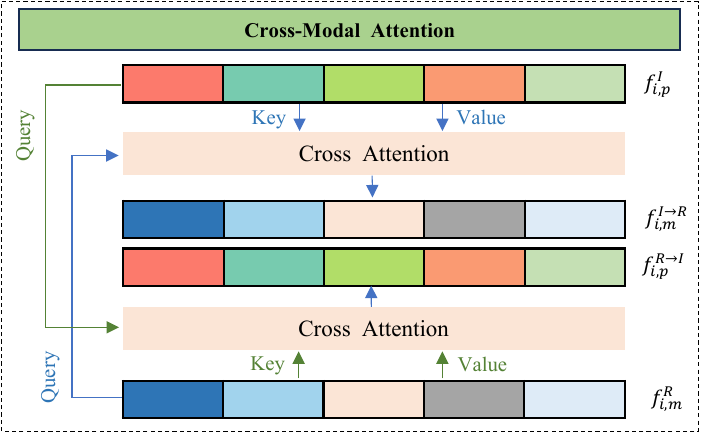}
\caption{Details of cross-modal attention module. We take $f_{i,p}^{I}$ as Query and $f_{i,m}^{R}$ as Key and Value, and compute the cross-modal image representation $f_{i,p}^{R\rightarrow I}$ through cross-attention. Similarly, by taking $f_{i,m}^{R}$ as Query and $f_{i,p}^{I}$ as Key and Value, the cross-modal text representation $f_{i,m}^{I\rightarrow R}$ can be also computed through cross-attention.}
\label{fig3:cross-attention}
\end{figure}

\subsection{Image and Report Representation}
For the paired image and text samples $(\bm{x}_{i}^{I}, \bm{x}_{i}^{R})$,  we extract features using an image encoder $E^I$ and a text encoder $E^R$. Specifically, we use ResNet50 \cite{he2016deep} as $E^I$ to encode $\bm{x}_{i}^{I}$ into representations of P sub-regions as local image representations, formulated as $\bm{f}_{i,p}^{I}\in \mathbb{R}^{P\times C_{I}}$, where $i$ denotes the $i$-th image, $p$ denotes the index of the $p$-th sub-region, and $C_I$ is the dimensionality of the image features. We obtain the global image feature $\bm{f} _{i}^{I}$ by performing self-attention pooling on $\bm{f}_{i,p}^{I}$.

Medical reports typically contain long sentences that require inference between words. Therefore, we adopt BioClinicalBERT \cite{alsentzer2019publicly}, pre-trained on the MIMIC III dataset\cite{johnson2016mimic}, as the text encoder $E^R$, which utilizes the self-attention mechanism to establish long-range semantic dependencies between words in medical reports. Each sentence in the medical image-text pairs usually describes one or several sub-regions in the image, indicating a correspondence between text sentences and image sub-regions, as shown in Fig. \ref{fig2:framework}. We extract sentence-level features from the text for alignment with the image sub-regions, unlike MGCA and GloRIA, which extract word-level features. 
Each text $\bm{x}_{i}^{R}$ is encoded into $M$ sentence-level representations, formulated as $\bm{f}_{i,m}^{R}\in  \mathbb{R}^{M\times C_{R}}$, where $m$ denotes the index of the $m$-th sentence in the $i$-th text, and $C_R$ represents the dimensionality of the text features. Note that $M$ varies across texts. Similar to the $\bm{f} _{i}^{I}$, we derive the global text feature $\bm{f} _{i}^{R}$ through self-attention pooling on the sentence-level representations $\bm{f}_{i,m}^{R}$ instead of a single [CLS] token. 

\subsection{Cross-Modal Representation Alignment} 
To perform cross-modal alignment, we use four Multi-Layer Perceptrons (MLPs) to map $\bm{f}_{i,p}^{I}$, $\bm{f}_{i,m}^{R}$ and $\bm{f} _{i}^{I}$, $\bm{f} _{i}^{R}$ to the same dimension $D$, facilitating cross-modal alignment. We treat the cross-modal representation corresponding to the current sample as the positive sample, while the other cross-modal representations within the same mini-batch are considered negative samples for computing contrastive loss.
\subsubsection{Global Alignment}
For multimodal learning, maximizing the mutual information between images and texts would be helpful to enhance the model's ability to understand the semantic relationships between different modalities. However, directly computing mutual information is often challenging because it requires precise joint and marginal probability distributions. To address this issue, we use the InfoNCE loss \cite{oord2018representation} to approximate the lower bound of mutual information through global contrastive learning, which allows for effective optimization without the need to explicitly calculate complex probability distributions.
Specifically, paired medical images and reports that have similar semantics after being mapped into a common feature space should be closer together, while non-matching pairs should be pushed further apart. 

To quantify this semantic relationship, we compute the cosine similarity between image sub-regions and text sentences:
\begin{equation}
\label{eq1}
\operatorname{sim}(\bm{x}_i^I,\bm{x}_i^R)=(\bar{\bm{f}}_i^I)^T\bar{\bm{f}}_i^R.
\end{equation}
Then, the global image-to-report alignment loss is defined as:
\begin{equation}
\label{eq2}
L_{g}^{I\rightarrow R}=-\frac{1}{B}\sum_{i=1}^{B}\log\frac{\exp(\operatorname{sim}(\bm{x}_{i}^{I},\bm{x}_{i}^{R})/\tau_1)}{\sum_{j=1}^{B}\exp(\operatorname{sim}(\bm{x}_{i}^{I},\bm{x}_{j}^{R})/\tau_1)},
\end{equation}
where $B$ denotes the batch size, and $\tau_1$ represents the temperature hyperparameter. 
Similarly, the global report-to-image alignment loss is given as:
\begin{equation}
\label{eq3}
L_{g}^{R\rightarrow I}=-\frac{1}{B}\sum_{i=1}^{B}\log\frac{\exp(\operatorname{sim}(\bm{x}_{i}^{R},\bm{x}_{i}^{I})/\tau_1)}{\sum_{j=1}^{B}\exp(\operatorname{sim}(\bm{x}_{i}^{R},\bm{x}_{j}^{I})/\tau_1)}.
\end{equation}
Finally, the total global loss, a combination of both the above losses, is as follows:
\begin{equation}
\label{eq4}
L_{global}=L_{g}^{I\rightarrow R} + L_{g}^{R\rightarrow I}.
\end{equation}

\subsubsection{Local Alignment with Clustering-Guided Negative Sampling}
In the medical domain, fine-grained information within images, such as tumor boundaries and lesion areas, is particularly crucial for dense prediction tasks like object detection and semantic segmentation. However, solely relying on global alignment may neglect these critical details due to its focus on the overall consistency between images and their textual descriptions. To address this limitation, a natural idea is to incorporate local alignment. It is worth noting that our implementation is toward cross-modal representations to exploit manual information between different modalities, which is more beneficial for capturing fine-grained details.

We use cross-attention, as shown in Fig. \ref{fig3:cross-attention}, to obtain the corresponding cross-modal representations $\bm{f}_{i,m}^{I\rightarrow R}$ and $\bm{f}_{i,p}^{R\rightarrow I}$ from the local features $\bm{f}_{i,m}^{R}$ and $\bm{f}_{i,p}^{I}$. For each sentence $m$, the cross-modal representation $\bm{f}_{i,m}^{I\rightarrow R}$ is obtained by letting $\bm{f}_{i,m}^{R}$ participate in the attention process with all paired image sub-region representations $\bm{f}_{i,p}^{I}$. Given that pathological information occupies only a small part of images or text, and not all sub-regions or sentences are informative, we use the sigmoid function instead of softmax for cross-attention. Softmax assumes equal contribution from all units, potentially diluting relevant information in noisy data. Sigmoid, which does not enforce a sum-to-one constraint, better suits multimodal medical data with sparse relevant features. Therefore, The local image-to-report cross-attention-based representation is expressed as:
\begin{equation}
\label{eq5}
\bm{f}_{i,m}^{I\rightarrow R}=\sum_{p=1}^{P}\sigma\left(\frac{\bm{Q}\bm{f}_{i,m}^{R}\cdot \bm{K}\bm{f}_{i,p}^{I}}{\sqrt{D}}\right)\cdot \bm{V}\bm{f}_{i,p}^{I},
\end{equation}
where $\bm{Q},\bm{K},\bm{V}\in \mathbb{R}^{D\times D}$ are learnable matrices, and $\sigma$ denotes the sigmoid function. Similarly, the formula for computing the report-to-image cross-attention-based representation is:
\begin{equation}
\label{eq6}
\bm{f}_{i,p}^{R\rightarrow I}=\sum_{m=1}^{M}\sigma\left(\frac{\bm{Q}\bm{f}_{i,p}^{I}\cdot \bm{K}\bm{f}_{i,m}^{R}}{\sqrt{D}}\right)\cdot \bm{V}\bm{f}_{i,m}^{R}.
\end{equation}

In contrastive learning, the quality of learned representations tends to improve as the number of high-quality negative samples increases. However, during local text alignment, due to the potential semantic similarity between sentences in different reports, simply sampling negative examples from all sentences in a mini-batch can lead to feature collapse, making it difficult for the model to distinguish between positive and negative sample pairs. To address the problem, some methods such as LOVT \cite{muller2022joint} and PRIOR \cite{cheng2023prior} restrict negative samples to sentence representations from the same report. However, it arises an additional issue that the model performance and generalization ability may be limited since the number of sentences in a single report is often small (sometimes only two or three sentences).

To tackle this issue, we propose the \textbf{C}ross-\textbf{M}odal \textbf{C}lustering-\textbf{G}uided \textbf{N}egative \textbf{S}ampling (\textbf{CM-CGNS}) method. Given that K-means clustering algorithm is renowned for its high computational efficiency and minimal parameter requirements -- making it particularly suitable for large-scale datasets -- CM-CGNS leverages a cross-modal K-means clustering mechanism to increase the number of negative samples essential for contrastive learning. For the resulting hard negative samples, we use grid search strategy to assign greater weight, thereby further improving the model learning efficiency and feature representation capabilities.

Specifically, we separately collect all sentences from the $l$-th mini-batch for both  $\bm{f}_{i,j}^{R}$ and corresponding $\bm{f}_{i,j}^{I\rightarrow R}$. The mini-batch contains a total of $M_l$ sentences, each of which is represented as $\bm{f}_{l,j}^{R}$ and $\bm{f}_{l,j}^{I\rightarrow R}$, where $j \in [1, M_l]$. We cluster all the sentence cross-modal representations $\bm{f}_{l,j}^{I\rightarrow R}$ into $K$ clusters $\bm{C}_l=\{\bm{C}_{l}^{1}, \bm{C}_{l}^{2}, ..., \bm{C}_{l}^{K}\}$, with cluster centers denoted as $\bm{c}_l=\{\bm{c}_{l}^{1}, \bm{c}_{l}^{2}, ..., \bm{c}_{l}^{K}\}$. Then, for each sentence representation $\bm{f}_{l,j}^{I\rightarrow R}$, we sort the clusters $\bm{C}_l$ and the cluster centers $\bm{c}_l$ in descending order based on the cosine similarity between $\bm{f}_{l,j}^{R}$ and each cluster center $\{\bm{c}_{l}^{1}, \bm{c}_{l}^{2}, ..., \bm{c}_{l}^{K}\}$, resulting in $\bm{C}_{l,j}=\{\bm{C}_{l,j}^{1}, \bm{C}_{l,j}^{2}, ..., \bm{C}_{l,j}^{K}\}$ and $\bm{c}_{l,j}=\{\bm{c}_{l,j}^{1}, \bm{c}_{l,j}^{2}, ..., \bm{c}_{l,j}^{K}\}$. 
At this point,
we select the cluster center $\bm{c}_{lj}^{2}$ (excluding $\bm{c}_{lj}^{1}$) that is similar to $\bm{f}_{l,j}^{R}$ as the hard negative for the sample $\bm{f}_{l,j}^{R}$.

When computing the image-to-report local alignment loss, we use the set of hard negatives $\bm{C}_{l,j}^*=\{\bm{c}_{l,1}^{2}, \bm{c}_{l,2}^{2}, ..., \bm{c}_{l,M_l}^{2}\}$ corresponding to each sentence $\bm{f}_{l,j}^{R}$ as additional negative samples, thereby increasing the number of negative samples and giving greater weight to each sample's corresponding hard negative. The formula for the image-to-report local alignment loss is defined as:
\begin{equation}
\label{eq7}
\scalebox{0.90}{
$
\begin{gathered}
L_{local\_report}^{CM-CGNS} = -\frac{1}{\sum_{l=1}^{B}M_{l}} \sum_{l=1}^{B} \sum_{j=1}^{M_{l}} \log \frac{\exp(\bm{f}_{l,j}^{R} \cdot (\bm{f}_{l,j}^{I\rightarrow R})^{T} / \tau_{3})}{\operatorname{H}(\bm{f}_{l,j}^{R},\bm{f}_{l,j}^{I\rightarrow R})},\\
\operatorname{H}(\bm{f}_{l,j}^{R},\bm{f}_{l,j}^{I\rightarrow R}) = \sum_{k=1}^{M_{l}} \bigl( \exp(\bm{f}_{l,j}^{R} \cdot (\bm{f}_{l,j}^{I\rightarrow R})^{T} / \tau_{3}) \\
\phantom{{}={}} + \bm{\mu} \exp(\bm{f}_{l,j}^{R} \cdot (\bm{C}_{l,j}^{*})^{T} / \tau_{3}) \bigr),
\end{gathered}
$}
\end{equation}
where $\tau_3$ is the temperature hyperparameter, $\bm{\mu}\in \mathbb{R}^{M_{B}\times M_{B}}$ is a weighting matrix, where the values on the diagonal represent the weight of the hard negatives. The optimal weight is determined through grid search.

False negatives are samples that should be classified as positive but are incorrectly labeled as negative. These samples are very close to positive samples in the feature space and should not be pushed away, as this could distort learned representations. In our experiments, we treat the samples in the cluster associated with the cluster center $\bm{c}_{l,j}^{1}$ that is most similar to the sentence representation $\bm{f}_{l,j}^{R}$ as false negatives $\bm{f}_{l,j}^{*}$ ($j\in [1, \operatorname{num}(\bm{C}_{l,j}^{1})]$), 
where $\operatorname{num}(\cdot)$ counts the number of samples in the cluster.
The similarity between the anchor and false negatives should be higher than with hard negatives but lower than with positives:
\begin{equation}
\label{eq8}
\cos(\bm{f}_{l,j}^{R},\bm{C}_{l,j}^{*})\leq\cos(\bm{f}_{l,j}^{R},\bm{f}_{l,j}^{*})\leq\cos(\bm{f}_{l,j}^{R},\bm{f}_{l,j}^{I\rightarrow R}).
\end{equation}
We use the bidirectional margin loss \cite{wang2023sncse} to compute the loss:
\begin{equation}
\label{eq9}
\begin{gathered}
\Delta x_{l,j}=\cos(\bm{f}_{l,j}^{R},\bm{f}_{l,j}^{*})-\cos(\bm{f}_{l,j}^{R},\bm{f}_{l,j}^{I\rightarrow R}),
\\L_{bml}=\operatorname{ReLU}(\Delta x_{l,j}+\alpha)+\operatorname{ReLU}(-\Delta x_{l,j}-\beta),
\end{gathered}
\end{equation}
where $\alpha$ and $\beta$ are two thresholds, $L_{bml}$ penalizes cases when $\Delta x_{l,j}$  is less than $-\alpha$ or greater than $-\beta$. This ensures that false negatives are neither too close to positive samples nor pushed too far away.

When computing the report-to-image local alignment loss, due to the potential similar information between adjacent image sub-regions, directly using these regions as negative samples could lead to feature collapse. Therefore, we adopt the SimSam \cite{chen2021exploring} alignment method, which constructs twin networks to learn image representations, reducing the dependency on the number of negative examples. The formula for the report-to-image local alignment loss is as follows:
\begin{equation}
\label{eq10}
\scalebox{0.90}{
$
\begin{aligned}L_{local\_image}^{SimSiam}&=-\frac{1}{BP}\sum_{i=1}^{B}\sum_{p=1}^{P}(\frac{1}{2}\operatorname{sim}(\operatorname{h}(\bm{f}_{i,p}^{I}),\operatorname{S}(\bm{f}_{i,p}^{R\rightarrow I}))\\&+\frac{1}{2}\operatorname{sim}(\operatorname{h}(\bm{f}_{i,p}^{R\rightarrow I}),\operatorname{S}(\bm{f}_{i,p}^{I}))),
\end{aligned}
$}
\end{equation}
where $\operatorname{h}(\cdot)$ denotes the MLP, $\operatorname{S}(\cdot)$ represents the stop-gradient operation, and 
$\operatorname{sim}(\cdot)$ computes the cosine similarity. Finally, the formula for the local alignment loss is:
\begin{equation}
\label{eq11}
L_{local}=L_{local\_report}^{CM-CGNS}+L_{bml}+L_{local\_image}^{SimSiam}.
\end{equation}

\subsection{Cross-Modal Masked Image Reconstruction} 
In downstream detection and segmentation tasks, the model needs to learn more accurate edge information of the object and finer pixel-level feature information. However, both global alignment and local alignment process the high-level image features extracted by the encoder, which could ignore the edge and detail features of the image, hindering the performance on downstream tasks. To address this issue, we designed a Cross-Modal Masked Image Reconstruction (CM-MIR) module to obtain low-level detail information of the image while enhancing the model's cross-modal information fusion capabilities. Specifically, for the local image representation $\bm{f}_{i,p}^{I}$, we randomly mask $\bm{f}_{i,p}^{I}$ with a probability of 50\% to get $\hat{\bm{f}}_{i,p}^{I}$. Then, we concatenate the cross-modal representation $\bm{f}_{i,p}^{R\rightarrow I}$ and the local image feature $\hat{\bm{f}}_{i,p}^{I}$ along the channel dimension, using $\bm{f}_{i,p}^{R\rightarrow I}$ to complete $\hat{\bm{f}}_{i,p}^{I}$. The fused image features are then passed through a decoder module consisting of five convolutional layers and bilinear interpolation for decoding image and reconstruction. We use L1 loss for cross-modal masked image reconstruction:
\begin{equation}
\label{eq12}
L_{re}=\frac{1}{B}\sum_{i=1}^{B}||\operatorname{head}(\operatorname{D_{I}}(\operatorname{F}(\bm{f}_{i,p}^{R\rightarrow I},\hat{\bm{f}}_{i,p}^{I})))-\bm{x}_{i}^{I}||_{1},
\end{equation}
where $\operatorname{F}(\cdot)$ denotes the feature fusion, $\operatorname{D_I}(\cdot)$ represents the image decoder, and $\operatorname{head}(\cdot)$ indicates the upsampling head.
\subsection{Overall Objective} 
The overall loss function of our proposed \textbf{CM-CGNS} consists of global loss, local loss, and cross-modal masked image reconstruction loss:
\begin{equation}
\label{eq13}
L=\lambda_{global}L_{global}+\lambda_{local}L_{local}+\lambda_{re}L_{re},
\end{equation}
where $\lambda_{global}, \lambda_{local}$ and $\lambda_{re}$ denote the hyperparameters.

\section{Experiments and Results}
\label{sec:experiments}
\subsection{Datasets}
\textbf{MIMIC-CXR 2.0.0}\cite{johnson2019mimic}. We pre-train our model on the MIMIC-CXR 2.0.0 dataset, which includes 377,110 JPG images and structured labels from 227,827 associated radiology reports. We follow \cite{zhang2022contrastive} to preprocess the dataset. Since the downstream datasets lack lateral view images, we retain only frontal-view chest images. We also extract only the impression and findings sections from the reports, discarding those that are empty or contain fewer than three tokens. Ultimately, this results in 232,286 image-text pairs.

\textbf{COVIDx}\cite{wang2020covid}. The COVIDx dataset comprises over 30,000 chest X-ray (CXR) images from more than 16,600 patients, including 16,490 COVID-19 positive images from over 2,800 patients. We utilize the original validation set as the test data and manually allocate 10\% of the original training set for validation.
\begin{table*}[htpb]
\centering
\small
\caption{Linear classification results by fine-tuning models on three datasets using $1\%$, $10\%$, and $100\%$ of the data, respectively. For the CheXpert and RSNA, we use the area under the ROC curve (AUC [$\%$]) as evaluation metric, while for the COVIDx, we use accuracy (ACC [$\%$]). The best and second-best results are highlighted in \textcolor{red}{red} and \textcolor{blue}{blue}, respectively.}
\resizebox{\linewidth}{!}{
\begin{tabular}{l|ccc|ccc|ccc}
\toprule
\multirow{2}{*}[-0.8ex]{Method}  & \multicolumn{3}{c|}{CheXpert (AUC)} & \multicolumn{3}{c|}{RSNA (AUC)}  & \multicolumn{3}{c}{COVIDx (ACC)} \\
\cmidrule{2-10}
 & 1\% & 10\% & 100\% & 1\% & 10\% & 100\% & 1\% & 10\% & 100\% \\
\midrule
Random Init.  & 55.37\textsubscript{\textpm 2.55}  & 63.55\textsubscript{\textpm 2.41}  & 66.74\textsubscript{\textpm 0.76}   &    74.70\textsubscript{\textpm 1.83} &  75.40\textsubscript{\textpm 0.77}   &   81.90\textsubscript{\textpm 0.58}   &   56.75\textsubscript{\textpm 3.20}  &  63.50\textsubscript{\textpm 0.88}  & 68.52\textsubscript{\textpm 0.43} \\
ImageNet Init.  & 66.13\textsubscript{\textpm 0.84}  & 68.58\textsubscript{\textpm 0.53}  &  74.93\textsubscript{\textpm 0.17}  &   79.66\textsubscript{\textpm 1.88}   &   84.39\textsubscript{\textpm 0.69}   &  85.93\textsubscript{\textpm 0.43}   &  66.25\textsubscript{\textpm 0.24}  &  78.50\textsubscript{\textpm 0.87}  &   84.25\textsubscript{\textpm 0.40} \\
\midrule
MoCo \cite{he2020momentum} & 68.35\textsubscript{\textpm 0.44}  & 77.75\textsubscript{\textpm 0.21}  & 84.57\textsubscript{\textpm 0.13}   &  80.61\textsubscript{\textpm 0.34}   &  84.36\textsubscript{\textpm 0.60}    &   85.20\textsubscript{\textpm 0.28}     &   65.69\textsubscript{\textpm 0.56}  &  76.88\textsubscript{\textpm 0.42}   &  84.17\textsubscript{\textpm 0.15} \\
MoCoV2 \cite{chen2020improved}   & 69.77\textsubscript{\textpm 0.53}  & 79.49\textsubscript{\textpm 0.26}   &  85.00\textsubscript{\textpm 0.11}   &   81.58\textsubscript{\textpm 0.60}    &   84.88\textsubscript{\textpm 0.67}    & 86.56\textsubscript{\textpm 0.26} &  66.70\textsubscript{\textpm 0.59}   &  78.12\textsubscript{\textpm 0.30}   &   85.73\textsubscript{\textpm 0.26} \\
SimCLR  \cite{chen2020simple}  & 67.40\textsubscript{\textpm 2.16}  & 78.45\textsubscript{\textpm 0.37} & 83.33\textsubscript{\textpm 0.18}  & 78.36\textsubscript{\textpm 2.07}  & 83.45\textsubscript{\textpm 0.56} &  84.73\textsubscript{\textpm 0.25}    &  64.10\textsubscript{\textpm 2.29}   &  77.21\textsubscript{\textpm 0.43}  &  84.67\textsubscript{\textpm 0.27} \\
\midrule
ConVIRT \cite{zhang2022contrastive}    & 77.93\textsubscript{\textpm 0.69}  &  81.77\textsubscript{\textpm 0.15} & 87.89\textsubscript{\textpm 0.38}   &   83.67\textsubscript{\textpm 0.84}  &  85.85\textsubscript{\textpm 0.51}  & 86.51\textsubscript{\textpm 0.17}  &   67.59\textsubscript{\textpm 0.31}  &  80.20\textsubscript{\textpm 0.56}  &  85.71\textsubscript{\textpm 0.43} \\
CLIP \cite{radford2021learning}  & 76.77\textsubscript{\textpm 0.26}  & 80.64\textsubscript{\textpm 0.53} & 85.21\textsubscript{\textpm 0.20}  &   82.02\textsubscript{\textpm 1.13}   & 85.70\textsubscript{\textpm 1.26}    & 86.24\textsubscript{\textpm 0.33} &     66.00\textsubscript{\textpm 1.17}   &   79.86\textsubscript{\textpm 0.42}    &  84.33\textsubscript{\textpm 0.49} \\
GLoRIA     \cite{huang2021gloria}   & 76.71\textsubscript{\textpm 1.47} & 81.50\textsubscript{\textpm 0.56}  & 88.23\textsubscript{\textpm 0.17}  &     83.41\textsubscript{\textpm 1.54}   &  85.47\textsubscript{\textpm 0.80}  & 87.24\textsubscript{\textpm 0.23}    &  67.93\textsubscript{\textpm 0.67} & 83.50\textsubscript{\textpm 0.45} &  87.67\textsubscript{\textpm 0.18} \\
LoVT \cite{muller2022joint}    & 78.26\textsubscript{\textpm 1.39}  & 82.65\textsubscript{\textpm 0.62}  & 87.94\textsubscript{\textpm 0.15}   &  84.43\textsubscript{\textpm 0.88}  &  86.00\textsubscript{\textpm 0.72}  & 88.14\textsubscript{\textpm 0.43}  &  68.82\textsubscript{\textpm 0.50}  &  83.06\textsubscript{\textpm 0.14}  &  87.45\textsubscript{\textpm 0.11} \\
PRIOR  \cite{cheng2023prior}   & 78.72\textsubscript{\textpm 1.33} & 81.81\textsubscript{\textpm 0.20} & 88.52\textsubscript{\textpm 0.18}  &  86.29\textsubscript{\textpm 0.41}  & 86.85\textsubscript{\textpm 0.18} & 88.33\textsubscript{\textpm 0.22} &  \textcolor{blue}{70.75\textsubscript{\textpm 0.73}}  &  84.50\textsubscript{\textpm 0.28} &  88.50\textsubscript{\textpm 0.33} \\
MGCA \cite{wang2022multi}  & \textcolor{blue}{81.63\textsubscript{\textpm 1.64}}  & \textcolor{blue}{87.00\textsubscript{\textpm 0.26}}  &  \textcolor{blue}{89.58\textsubscript{\textpm 0.18}}  & \textcolor{blue}{87.18\textsubscript{\textpm 0.36}}  & \textcolor{blue}{87.62\textsubscript{\textpm 0.17}}  & \textcolor{blue}{88.93\textsubscript{\textpm 0.11}}  &  69.70\textsubscript{\textpm 0.34} &  \textcolor{blue}{87.00\textsubscript{\textpm 0.31}}  & \textcolor{blue}{91.00\textsubscript{\textpm 1.23}} \\
\midrule
CM-CGNS (Ours) &\textcolor{red}{82.05\textsubscript{\textpm 1.53}} & \textcolor{red}{87.24\textsubscript{\textpm 0.38}} & \textcolor{red}{89.86\textsubscript{\textpm 0.19}} & \textcolor{red}{87.32\textsubscript{\textpm 0.08}} & \textcolor{red}{88.55\textsubscript{\textpm 0.15}}  &\textcolor{red}{89.41\textsubscript{\textpm 0.17}} & \textcolor{red}{75.21\textsubscript{\textpm 0.51}}  & \textcolor{red}{88.50\textsubscript{\textpm 0.26}} &  \textcolor{red}{92.25\textsubscript{\textpm 0.21}} \\
\bottomrule
\end{tabular}
}
\label{tab:supervised-cls}
\end{table*}

\textbf{CheXpert}\cite{irvin2019chexpert}. We use CheXpert as our classification dataset. The CheXpert dataset is a large chest X-ray dataset containing 224,316 images from 65,240 patients. We removed the lateral view images, resulting in 191,229 frontal view images. Each image is labeled for five independent diseases, and the task is to classify each image into five individual binary labels: atelectasis, cardiomegaly, consolidation, edema, and pleural effusion. We use the expert-labeled validation set as our test set.

\textbf{RSNA Pneumonia Detection}\cite{shih2019augmenting}. We utilize the second stage of the RSNA Pneumonia dataset for classification, object detection, and semantic segmentation tasks. This dataset comprises approximately 29,700 frontal chest X-ray images. For the classification experiments, we divide the dataset into training, validation, and test sets in a 70\%/15\%/15\% ratio. In the object detection and semantic segmentation experiments, the dataset is split into training, validation, and test sets with a 60\%/20\%/20\% ratio.

\textbf{Object CXR}\cite{healthcare2020object}. Object CXR contains 9,000 frontal chest X-rays with detection targets for foreign objects. We split the original training set into a training set and a validation set, containing 6,400 images and 1,600 images, respectively. The original development set is used as the test set, which contains 1,000 images.

\textbf{SIIM-ACR Pneumothorax}\cite{siim-acr-pneumothorax-segmentation}. The SIIM-ACR Pneumothorax dataset contains 12,047 chest radiographs, each with manually annotated segmentation masks for pneumothorax. We split the dataset into training, validation, and test sets with a 70\%/15\%/15\% ratio.

\subsection{Implementation Details} 

We use BioClinicalBERT\cite{alsentzer2019publicly} as the text encoder and ResNet-50 \cite{he2016deep} as the image encoder. We resize the images to 224×224, set the dimension of the latent space to 768, and set the temperature hyperparameters $\tau_1$, $\tau_2$ and $\tau_3$ to 0.01. The loss weight coefficients are set as $\lambda_{global} = \lambda_{local} = 1.0$ and $\lambda_{re} = 10.0$. We performed a grid search over the weight $\mu$, as well as thresholds $\alpha$ and $\beta$, finding that $\mu = 3.0 $, $\alpha = 0.2 $, and $\beta = 0.5 $ yield the optimal performance. For the pre-training stage, the batch size is set to 64, and the initial learning rate is 1e-5 with a cosine decay strategy. The first stage is set to 10 epochs, and the second stage is set to 20 epochs. We conduct our pre-training experiments on 4 NVIDIA RTX 3090 GPUs.

We compare our proposed method with other widely-used self-supervised methods, including single-modal image SSL methods (MoCo, MoCoV2, SimCLR) and multimodal VLP methods (ConVIRT, CLIP, GloRIA, LoVT, MGCA, PRIOR). Additionally, we include a randomly initialized ResNet50 model and a ResNet50 model pre-trained on the ImageNet dataset for comparison. In the single-modal image self-supervised learning (SSL) methods, only the image modality data is used, constructing positive and negative sample pairs through data augmentation for contrastive learning pre-training. Among the multimodal VLP methods, ConVIRT and CLIP do not feature local alignment modules that would facilitate the extraction of fine-grained pathological information. On the other hand, GloRIA, LoVT, MGCA, and PRIOR have specifically designed their own local alignment modules to extract local pathological information, making them more suitable for medical image-text SSL. 
Since some of the datasets used for pre-training are not MIMIC-CXR, for a fair comparison, we reproduce all the methods on the MIMIC-CXR dataset using their respective official code and default parameter settings, and then use the obtained pre-trained models for downstream tasks.

\begin{table}[htpb]
    \centering 
    \caption{Object detection results (mAP [$\%$]) on RSNA and Object CXR datasets, fine-tuned using $1\%, 10\%, 100\%$ of the training data. "-" indicates that the mAP is less than $1\%$.}
    \resizebox{\linewidth}{!}{
    \begin{tabular}{l|ccc|ccc}
    \toprule
    \multirow{2}{*}[-0.8ex]{Method} & \multicolumn{3}{c|}{RSNA} & \multicolumn{3}{c}{Object CXR}\\
    \cmidrule{2-7}
     & 1\% & 10\% & 100\% & 1\% & 10\% & 100\% \\
    \midrule
    Random Init. & - & 6.07\textsubscript{\textpm 1.56} & 10.39\textsubscript{\textpm 1.63} & - & 2.89\textsubscript{\textpm 1.92} & 4.51\textsubscript{\textpm 1.57} \\
    ImageNet Init. & 3.99\textsubscript{\textpm 3.24} & 14.47\textsubscript{\textpm 2.77} & 20.78\textsubscript{\textpm 1.98} & - & 7.90\textsubscript{\textpm 2.90} & 14.30\textsubscript{\textpm 2.21} \\ 
    \midrule
    MoCo \cite{he2020momentum} & 13.67\textsubscript{\textpm 2.60} & 16.48\textsubscript{\textpm 2.34} & 23.60\textsubscript{\textpm 1.90} & - & 6.40\textsubscript{\textpm 2.43} & 14.31\textsubscript{\textpm 1.88} \\
    MoCoV2 \cite{chen2020improved} & 16.82\textsubscript{\textpm 2.85} & 18.33\textsubscript{\textpm 2.16} & 24.25\textsubscript{\textpm 1.79} & - & 7.93\textsubscript{\textpm 2.15} & 15.62\textsubscript{\textpm 1.97} \\
    SimCLR \cite{chen2020simple} & 14.33\textsubscript{\textpm 2.53} & 16.90\textsubscript{\textpm 2.10} & 23.81\textsubscript{\textpm 1.97} & - & 6.94\textsubscript{\textpm 1.94} & 14.40\textsubscript{\textpm 1.83} \\
    \midrule
    ConVIRT \cite{zhang2022contrastive} & 16.21\textsubscript{\textpm 1.86} & 17.34\textsubscript{\textpm 1.88} & 22.63\textsubscript{\textpm 1.43} & - & 7.84\textsubscript{\textpm 2.08} & 15.86\textsubscript{\textpm 1.77} \\
    CLIP \cite{radford2021learning} & 14.98\textsubscript{\textpm 2.53} & 17.02\textsubscript{\textpm 2.31} & 22.17\textsubscript{\textpm 2.17} & - & 7.47\textsubscript{\textpm 1.93} & 15.36\textsubscript{\textpm 2.10} \\
    GLoRIA \cite{huang2021gloria} & 16.35\textsubscript{\textpm 2.08} & 18.70\textsubscript{\textpm 1.75} & 23.18\textsubscript{\textpm 2.87} & - & 8.73\textsubscript{\textpm 2.16} & 16.76\textsubscript{\textpm 1.94}  \\
    LoVT\cite{muller2022joint} & 15.91\textsubscript{\textpm 2.64} & 18.60\textsubscript{\textpm 2.21} & 24.54\textsubscript{\textpm 1.26} & - & 8.42\textsubscript{\textpm 2.51} & 16.51\textsubscript{\textpm 1.33} \\
    PRIOR  \cite{cheng2023prior} & 16.34\textsubscript{\textpm 1.97} & \textcolor{blue}{19.09\textsubscript{\textpm 1.92}} & 24.65\textsubscript{\textpm 1.18} & - & 10.18\textsubscript{\textpm 2.17} & 16.63\textsubscript{\textpm 1.62}  \\
    MGCA \cite{wang2022multi} & \textcolor{blue}{17.76\textsubscript{\textpm 2.36}} & 18.61\textsubscript{\textpm 1.86} & \textcolor{blue}{26.42\textsubscript{\textpm 1.23}} & - & \textcolor{red}{12.67\textsubscript{\textpm 2.21}} & \textcolor{blue}{17.43\textsubscript{\textpm 1.96}}  \\ 
    \midrule
    CM-CGNS (Ours) & \textcolor{red}{19.17\textsubscript{\textpm 2.46}} & \textcolor{red}{20.54\textsubscript{\textpm 1.95}} & \textcolor{red}{27.79\textsubscript{\textpm 1.16}} & - & \textcolor{blue}{10.24\textsubscript{\textpm 2.18}} & \textcolor{red}{18.71\textsubscript{\textpm 1.71}}  \\
    \bottomrule
    \end{tabular}
    }
    \label{tab:exp_det}
\end{table}

\subsection{Results on Medical Image Classification}
We conducted medical image classification experiments on the CheXpert, RSNA, and COVIDx datasets. Following previous work \cite{wang2022multi}, we added a linear classification head to the image encoder obtained from the pre-training stage. During the downstream tasks, we froze the weight parameters of the image encoder and only trained the linear classification head to evaluate the transfer performance of the pre-trained image encoder. Additionally, we evaluated the performance differences of the model when fine-tuned using 1\%, 10\%, and 100\% of the downstream data, further validating the data efficiency of our model. The experimental results of the classification are shown in Table \ref{tab:supervised-cls}.

As shown in Table \ref{tab:supervised-cls}, all methods utilize ResNet50 as the backbone network. Among these, methods employing image SSL outperform those using random initialization and ImageNet initialization. Furthermore, methods that leverage VLP surpass those using image SSL. Additionally, methods that incorporate a specifically designed local alignment module significantly outperform standard VLP methods. Our CM-CGNS achieves the best results across nine different settings on three datasets, outperforming the current state-of-the-art methods, MGCA and PRIOR. This demonstrates the effectiveness of our strategy to enhance contrastive learning by increasing the number of negative samples. 

\subsection{Results on Medical Object Detection}
We conduct object detection experiments on the RSNA and Object CXR datasets. For a fair comparison, all methods adopt a combined architecture of ResNet50 and YOLOv3, with consistent parameter configurations. The experimental results are shown in Table \ref{tab:exp_det}. We observe that the metrics for methods using VLP are generally better than those using only the image modality for pretraining. However, some SSL methods that use only the image modality, such as MoCoV2, outperform certain VLP methods like ConVIRT and CLIP, which do not incorporate a local alignment module. This may be because, when MIMIC-CXR dataset is used for pretraining, the medical reports may not provide detailed annotations of the specific locations and types of pneumonia. As a result, VLP methods face challenges in extracting and utilizing this information. In contrast, SSL methods primarily rely on the information within the images themselves, making them less affected by the imprecision of the reports, thus performing better on the RSNA dataset. However, our method achieves the best results in five out of six different experimental settings, demonstrating that our model can effectively extract features even when dealing with complex datasets.

\begin{table}[htpb]
    \centering
    \caption{Semantic segmentation results (Dice [$\%$]) on SIIM and RSNA datasets, fine-tuned with $1\%, 10\%, 100\%$ training data.}
   \resizebox{\linewidth}{!}{
    \begin{tabular}{l| c c c |c c c}
    \toprule
   \multirow{2}{*}[-0.8ex]{Method} & \multicolumn{3}{c|}{SIIM} & \multicolumn{3}{c}{RSNA}\\
   \cmidrule{2-7}
    & 1\% & 10\% & 100\% & 1\% & 10\% & 100\% \\
    \midrule
    Random Init. & 5.83\textsubscript{\textpm 1.78} & 20.75\textsubscript{\textpm 2.23} & 39.74\textsubscript{\textpm 0.69} & 47.78\textsubscript{\textpm 1.83} & 54.01\textsubscript{\textpm 1.24} & 62.06\textsubscript{\textpm 0.76} \\
    ImageNet Init. & 14.72\textsubscript{\textpm 3.23} & 50.75\textsubscript{\textpm 2.98} & 53.22\textsubscript{\textpm 1.67} & 56.16\textsubscript{\textpm 2.61} & 60.38\textsubscript{\textpm 2.13} & 67.99\textsubscript{\textpm 1.79} \\ 
    \midrule
    MoCo \cite{he2020momentum} & 23.77\textsubscript{\textpm 1.47} & 52.70\textsubscript{\textpm 1.52} & 58.60\textsubscript{\textpm 1.44} & 62.80\textsubscript{\textpm 1.67} & 63.77\textsubscript{\textpm 0.83} & 65.46\textsubscript{\textpm 0.89} \\
    MoCoV2 \cite{chen2020improved} & 25.67\textsubscript{\textpm 2.98} & 53.91\textsubscript{\textpm 2.76} & 62.86\textsubscript{\textpm 1.75} & 63.55\textsubscript{\textpm 2.37} & 64.86\textsubscript{\textpm 1.95} & 67.40\textsubscript{\textpm 1.36} \\
    SimCLR \cite{chen2020simple} & 22.25\textsubscript{\textpm 1.85} & 47.73\textsubscript{\textpm 0.98} & 57.37\textsubscript{\textpm 0.75} & 58.64\textsubscript{\textpm 1.18} & 61.57\textsubscript{\textpm 0.92} & 64.27\textsubscript{\textpm 0.77} \\ 
    \midrule
    ConVIRT \cite{zhang2022contrastive} & 25.20\textsubscript{\textpm 2.45} & 52.64\textsubscript{\textpm 1.51} & 59.89\textsubscript{\textpm 0.96} & 64.03\textsubscript{\textpm 1.86} & 65.00\textsubscript{\textpm 1.43} & 67.38\textsubscript{\textpm 1.12} \\
    CLIP \cite{radford2021learning} & 24.41\textsubscript{\textpm 1.98} & 51.19\textsubscript{\textpm 2.42} & 58.20\textsubscript{\textpm 1.81} & 62.34\textsubscript{\textpm 1.26} & 64.57\textsubscript{\textpm 2.33} & 65.82\textsubscript{\textpm 1.42} \\
    GLoRIA \cite{huang2021gloria} & 26.17\textsubscript{\textpm 2.68} & 57.67\textsubscript{\textpm 2.13} & 62.39\textsubscript{\textpm 1.67} & 66.74\textsubscript{\textpm 2.48} & 68.20\textsubscript{\textpm 1.36} & 68.94\textsubscript{\textpm 1.03}  \\
    LoVT\cite{muller2022joint} & 25.64\textsubscript{\textpm 1.16} & 54.21\textsubscript{\textpm 1.67} & 61.28\textsubscript{\textpm 1.28} & 65.56\textsubscript{\textpm 0.74} & 67.75\textsubscript{\textpm 0.96} & 68.19\textsubscript{\textpm 0.81} \\ 
    PRIOR  \cite{cheng2023prior} & 27.00\textsubscript{\textpm 2.13} & 57.32\textsubscript{\textpm 1.77} & \textcolor{blue}{66.19\textsubscript{\textpm 1.26}} & 67.92\textsubscript{\textpm 1.37} & 66.79\textsubscript{\textpm 1.18} & 71.30\textsubscript{\textpm 0.89}  \\
    MGCA \cite{wang2022multi} & \textcolor{red}{35.11\textsubscript{\textpm 2.46}} & \textcolor{blue}{57.86\textsubscript{\textpm 1.93}} & 63.41\textsubscript{\textpm 1.46} & \textcolor{red}{68.66\textsubscript{\textpm 1.02}} & \textcolor{blue}{69.53\textsubscript{\textpm 1.24}} & \textcolor{blue}{69.97\textsubscript{\textpm 0.97}}  \\ 
    \midrule
    CM-CGNS (Ours) & \textcolor{blue}{31.97\textsubscript{\textpm 2.22}} & \textcolor{red}{58.42\textsubscript{\textpm 1.73}} & \textcolor{red}{67.41\textsubscript{\textpm 1.22}} & \textcolor{blue}{68.52\textsubscript{\textpm 1.14}} & \textcolor{red}{72.21\textsubscript{\textpm 1.06}} & \textcolor{red}{72.56\textsubscript{\textpm 0.82}}  \\
    \bottomrule
    \end{tabular}
   }
    \label{tab:exp_seg}
\end{table}

\subsection{Results on Medical Semantic Segmentation}
We conduct semantic segmentation experiments on the RSNA and SIIM datasets, with all methods utilizing the same architecture that combines ResNet50 with U-Net. From Table \ref{tab:exp_seg}, it is evident that methods incorporating a specifically designed local alignment module exhibit significantly higher performance compared to those lacking such a module. This observation underscores the critical importance of local detail information in medical image segmentation. Furthermore, the methods leveraging VLP consistently outperform those relying solely on image modality for pre-training. This suggests that integrating information from both image and text modalities during training enhances the model's capability to extract meaningful image representations. Notably, among all evaluated approaches, our proposed CM-CGNS method stands out, achieving top performance in four out of six experimental settings and second place in the remaining two settings. These results affirm the effectiveness of the CM-CGNS module and the cross-modal masked image reconstruction mechanism in enhancing the extraction of local image features, which is particularly beneficial for medical image segmentation tasks.

\begin{figure}
\centering
\includegraphics[width=0.95\linewidth, keepaspectratio]{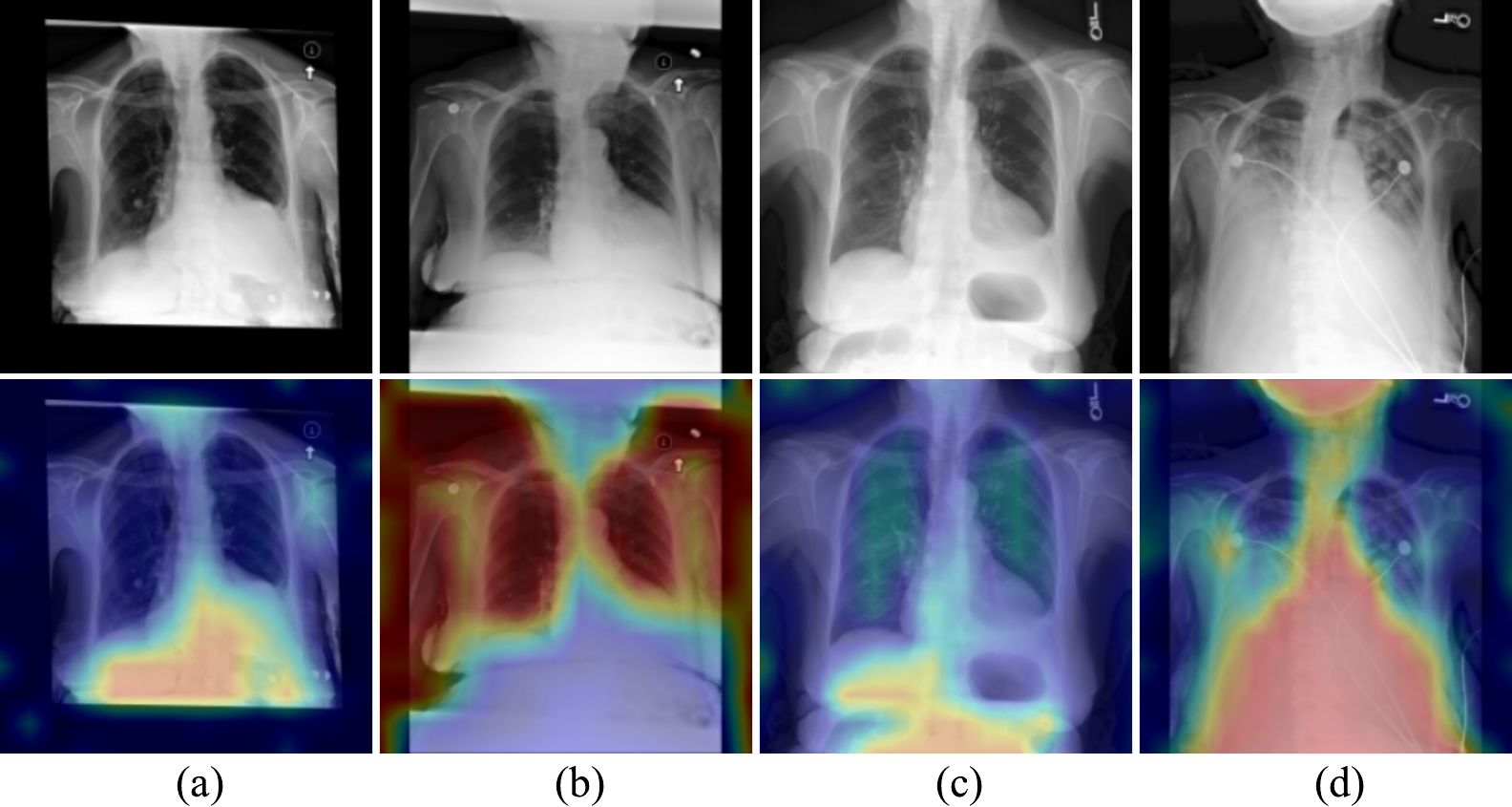}
\caption{Visualization of Cross-Modal Attention Maps. The sentences corresponding to the four images are: (a) ``persistent small bilateral effusions larger on the left which have decreased in size." (b) ``no pneumothorax is seen." (c) ``enlarging moderate left pleural effusion." (d) ``mild to large bilateral right greater than left pleural effusions."}
\label{fig4:attention_map}
\end{figure}

\subsection{Visualization and Ablation Studies}
\begin{figure}
\centering
\includegraphics[width=0.9\linewidth, keepaspectratio]{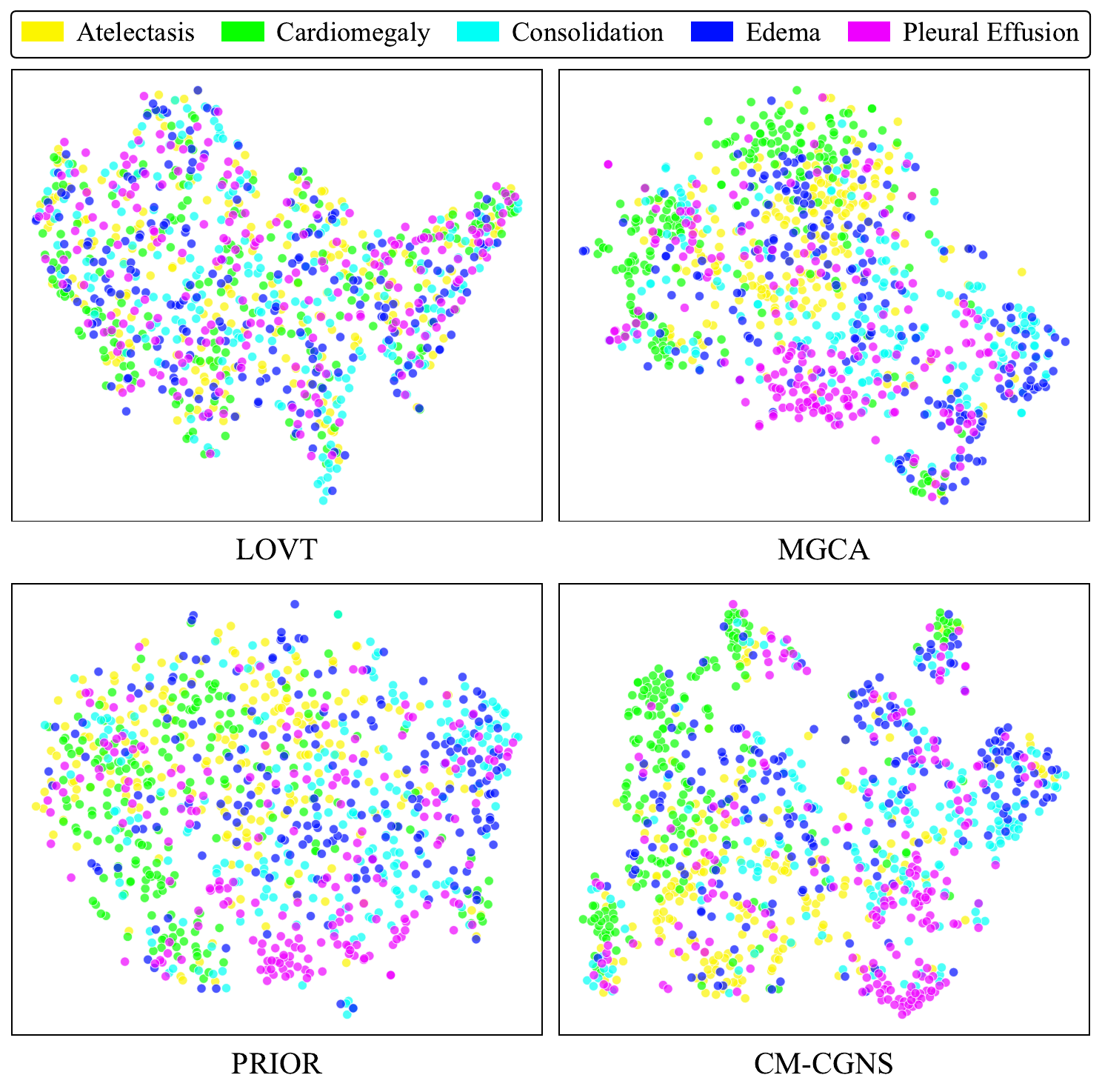}
\caption{t-SNE visualization of the representations from the last layer of the image encoder. Five ground truth labels from the CheXpert 5x200 dataset are shown.}
\label{fig5:t-SNE}
\end{figure}

We visualize report-to-image cross-modal attention maps, as shown in Fig. \ref{fig4:attention_map}. It can be seen that our model is capable of aligning sentence representations with their corresponding image regions, which aids in obtaining more fine-grained features and thus enhances the model representation capability. To further evaluate the capability, we employ t-SNE \cite{JMLR:v9:vandermaaten08a} to visualize the embeddings extracted from the final layer of our image encoder on CheXpert 5$\times$200 dataset, as shown in Fig. \ref{fig5:t-SNE}. Our findings reveal that, within the t-SNE plots, the points corresponding to different diseases in our method are distinctly separable, showcasing a more pronounced clustering effect. Conversely, the top-performing alternative approaches tend to generate feature distributions that are more isotropic and homogeneous, thereby struggling to effectively distinguish between various disease categories.

\begin{table}[htpb]
  \centering
  \caption{Ablation study results in terms of CM-CGNS and CM-MIR on CheXpert and SIIM. The best experimental results are highlighted in \textbf{bold}}
  \resizebox{\linewidth}{!}{
  \begin{tabular}{c c | c c c | c c c}
    \toprule
    \multicolumn{2}{c|}{Method} & \multicolumn{3}{c|}{CheXpert (AUC)}  & \multicolumn{3}{c}{SIIM (Dice)}\\
    \midrule
    CM-CGNS & CM-MIR & 1\% & 10\% & 100\% & 1\% & 10\% & 100\% \\
    \midrule
     & & 77.21\textsubscript{\textpm 1.82} & 81.47\textsubscript{\textpm 0.63} & 86.30\textsubscript{\textpm 0.56} & 26.86\textsubscript{\textpm 2.47} & 51.74\textsubscript{\textpm 2.20} & 58.97\textsubscript{\textpm 1.46} \\
     \checkmark & & 79.68\textsubscript{\textpm 1.77} & 84.36\textsubscript{\textpm 0.47} & 87.71\textsubscript{\textpm 0.31} & 28.70\textsubscript{\textpm 2.26} & 54.84\textsubscript{\textpm 1.94} & 63.12\textsubscript{\textpm 1.16} \\
     & \checkmark & 79.21\textsubscript{\textpm 1.60} & 83.70\textsubscript{\textpm 0.62} & 87.38\textsubscript{\textpm 0.36} & 28.72\textsubscript{\textpm 2.39} & 53.99\textsubscript{\textpm 2.10} & 62.82\textsubscript{\textpm 1.25} \\
     \checkmark & \checkmark & \textbf{82.05\textsubscript{\textpm 1.53}} & \textbf{87.24\textsubscript{\textpm 0.38}} & \textbf{90.04\textsubscript{\textpm 0.19}} & \textbf{31.97\textsubscript{\textpm 2.22}} & \textbf{58.42\textsubscript{\textpm 1.73}} & \textbf{67.41\textsubscript{\textpm 1.22}} \\
    \bottomrule
    \end{tabular}
  }
  \label{Tab:ablation}
\end{table}

In the ablation studies, the baseline method consists solely of the global alignment module and a conventional local alignment module. We add our proposed CM-CGNS and CM-MIR methods to the baseline model individually to verify the effectiveness of each component. As shown in Table \ref{Tab:ablation}, when any one of these methods is added to the baseline, there is an improvement in the experimental metrics, indicating that all the proposed methods are effective. Notably, The CM-CGNS module contributes more to the performance improvement, demonstrating its capability to effectively align the semantics between different modalities and retain important fine-grained information within images. When two methods are combined, there is a further performance improvement, suggesting that the methods complement each other.

\section{Conclusion}
\label{sec:conclusion}
In this paper, we introduce a Cross-Modal Clustering-Guided Negative Sampling (CM-CGNS) method for self-supervised joint learning from medical images and reports. CM-CGNS increases the number of negative samples in contrastive learning through cross-modal clustering negative sampling and enhances the model learning efficiency and representation capability by assigning greater weight to hard negative samples. Additionally, we design a Cross-Modal Masked Image Reconstruction (CM-MIR) module that utilizes local text-to-image features obtained through cross-attention to complete masked local image features. CM-MIR can improve the model's cross-modal information interaction capabilities and preserve low-level image features beneficial for downstream fine-grained tasks. Extensive experimental results on downstream tasks such as classification, detection, and segmentation demonstrate that our proposed method exhibits excellent learning efficiency and image representation transfer capability. Despite its strengths, our work still encounters certain limitations. One of the main concerns is that the computational complexity associated with clustering could become a bottleneck when processing large-scale datasets. In future work, we aim to optimize the clustering algorithms to better adapt to large-scale dataset scenarios, which is crucial for expanding the applicability and scalability of our approach.

\bibliographystyle{IEEEtran}
\bibliography{IEEEabrv,mybib}
\end{document}